%% file: main.tex
\documentclass[10pt,twocolumn,letterpaper]{article}

\usepackage[pagenumbers]{cvpr} 

\input{preamble}
\definecolor{cvprblue}{rgb}{0.21,0.49,0.74}
\usepackage[pagebackref,breaklinks,colorlinks,allcolors=cvprblue]{hyperref}
\usepackage[accsupp]{axessibility}  

\def\paperID{21640} 
\def\confName{CVPR}
\def\confYear{2026}

\title{GeoSURGE: Geo-localization using  Semantic Fusion with \\ Hierarchy of Geographic Embeddings}

\author{Angel Daruna, Nicholas Meegan, Han-Pang Chiu, Supun Samarasekera, Rakesh Kumar\\
SRI International\\
Princeton, NJ, USA\\
{\tt\small \{angel.daruna,nicholas.meegan,han-pang.chiu,supun.samarasekera,rakesh.kumar\}@sri.com}
}

\begin{document}
\maketitle
\input{sec/0_abstract}    
\input{sec/1_intro}
\input{sec/2_related}
\input{sec/3_approach}
\input{sec/4_experiments}
\input{sec/5_conclusion}

{
    \small
    \bibliographystyle{ieeenat_fullname}
    \bibliography{main}
}

\input{sec/X_suppl}

\end{document}

%% file: sec/0_abstract.tex
\begin{abstract}
Worldwide visual geo-localization aims to determine the geographic location of an image anywhere on Earth using only its visual content. Despite recent progress, learning expressive representations of geographic space remains challenging due to the inherently low-dimensional nature of geographic coordinates. We formulate global geo-localization as aligning the visual representation of a query image with a learned geographic representation. Our approach explicitly models the world as a hierarchy of learned geographic embeddings, enabling a distributed and multi-scale representation of geographic space. In addition, we introduce a semantic fusion module that efficiently integrates appearance features with semantic segmentation through latent cross-attention, producing a more robust visual representation for localization. Experiments on five widely used geo-localization benchmarks demonstrate that our method achieves new state-of-the-art results on 22 of 25 reported metrics. Ablation studies show that these improvements are primarily driven by the proposed geographic representation and semantic fusion mechanism.
\end{abstract}

%% file: sec/1_intro.tex
\section{Introduction}
\label{sec:intro}

Visual geo-localization supports important applications in autonomous systems, emergency response, and personalized digital services by enabling machines to estimate the geographic location of an image. Several variants of the visual geo-localization problem have been studied, including city-scale \cite{masone2021survey}, cross-view \cite{lin2013cross}, and global \cite{astruc2024openstreetview}. Global visual geo-localization remains challenging due to the immense diversity of scenes across the world. In this work, we focus on the global geo-localization problem where the location of a single query image must be estimated without additional contextual information.

Most prior global geo-localization approaches fall into two main categories: retrieval-based methods \cite{hays2008im2gps,hays2015large} and classification-based methods \cite{weyand2016planet}. Retrieval-based approaches estimate location by comparing a query image against a large reference database of geotagged images. In contrast, classification-based approaches discretize the Earth's surface into geographic cells (geocells) and train a classifier to predict the cell containing the query image. Each paradigm has important limitations. Retrieval-based methods require expensive large-scale similarity search at inference time, while classification-based approaches must balance the trade-off between spatial resolution and global coverage when defining geocells.

Recent state-of-the-art methods include GeoCLIP \cite{vivanco2024geoclip}, Img2Loc \cite{zhou2024img2loc}, and G3 \cite{jia2024g3}. GeoCLIP replaced image references with GPS coordinates and introduced specialized components such as Random Fourier Features to mitigate the information loss incurred when representing geographic locations in low-dimensional space. Img2Loc leveraged large vision-language models (LVLMs) together with geotagged image references to significantly improve localization performance. G3 subsequently combined ideas from both GeoCLIP and Img2Loc. Despite this progress, learning expressive representations of geography remains challenging due to the inherently low dimensions of GPS.

In this work, we introduce GeoSURGE (\underline{Geo}-localization using \underline{S}emantic F\underline{u}sion with Hie\underline{r}archy of \underline{G}eographic \underline{E}mbeddings), a method that bridges retrieval-based and classification-based approaches. GeoSURGE represents the world as a hierarchy of geographic embeddings, forming a distributed and multi-scale representation of geographic space. Similar to classification-based methods, the Earth's surface is partitioned into geocells. However, instead of treating geocells as discrete class labels, GeoSURGE learns a feature embedding for each geocell. Global geo-localization is then formulated as matching the visual representation of a query image with the learned geographic representations.

\begin{figure*} [t]
  \fontsize{10}{10}\selectfont
  \centering
   \includegraphics[width=0.95\linewidth]{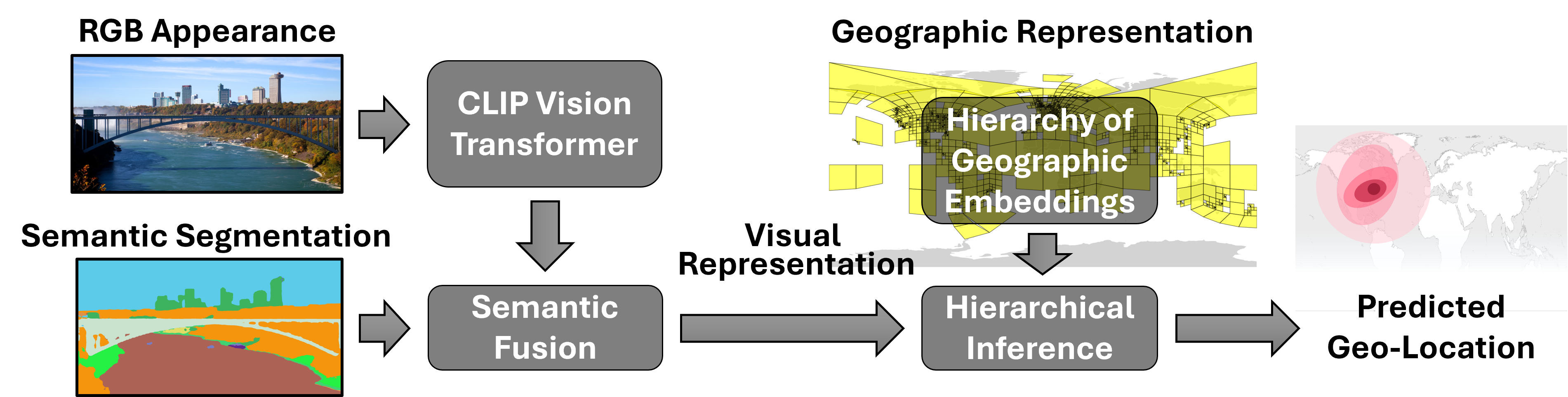}
   \caption{GeoSURGE Approach Overview: The location of an input image is predicted via hierarchical inference, by matching the visual representation of the image against the geographic representation, which is learned beforehand. The visual representation is generated from the semantic fusion module, which enriches appearance features with semantic segmentation.}
   \vspace{-0.5cm}
   \label{fig:overview}
\end{figure*}

In addition, GeoSURGE introduces a semantic fusion module that enriches visual representations by integrating appearance and semantic information. Specifically, CLIP features extracted from the RGB image \cite{haas2024pigeon,vivanco2024geoclip} are fused with features derived from a semantic segmentation map using latent cross-attention. Appearance-based features capture fine-grained visual cues, while semantic segmentation provides more invariant scene structure that is robust to changes in illumination, weather, and viewpoint. Furthermore, semantic cues help identify regions that are unreliable for localization, such as humans or vehicles.

We evaluate GeoSURGE on five widely used geo-localization benchmarks: IM2GPS \cite{hays2008im2gps}, IM2GPS3k \cite{vo2017revisiting}, YFCC4k \cite{vo2017revisiting}, YFCC26k \cite{muller2018geolocation}, and GWS15k \cite{clark2023we}. Across these datasets, GeoSURGE achieves  state-of-the-ar results on 22 out of 25 reported metrics. We further conduct extensive ablation studies to analyze the impact of each component of the system. These experiments show that the improvements are primarily driven by the proposed geographic embedding hierarchy and the semantic fusion representation. Overall, our results demonstrate that task-specific geo-localization architectures remain highly effective for improving global geo-localization accuracy.

In summary, GeoSURGE makes the following contributions to planet-scale image geo-localization:
\begin{enumerate}
    \item We introduce a hierarchical geographic embedding representation that models Earth's surface as a distributed feature hierarchy rather than discrete geographic classes.
    \item We propose a fusion module that integrates semantic segmentation and appearance features using latent cross-attention to produce enriched visual representations.
    \item We achieve new state-of-the-art performance on 22 out of 25 benchmark metrics across five widely used geo-localization datasets.
\end{enumerate}


%% file: sec/2_related.tex
\section{Related Works}
\label{sec:related}

Planet-scale image geo-localization approaches can be mainly categorized into two types: retrieval-based and classification-based methods, with some works combining the two approaches for improved performance.

\textbf{Classification-Based Methods} divide the world into geocells and determine which geocell contains a query image to infer its location \cite{weyand2016planet}. The seminal work of PlaNet partitioned Earth into geocells with respect to the distribution of training images using Google's S2 library and trained a convolutional neural network to assign a query image to a geocell \cite{weyand2016planet}. Many later works explored other partitioning methods including hierarchies \cite{seo2018cplanet,clark2023we, vo2017revisiting, muller2018geolocation, izbicki2020exploiting, astruc2024openstreetview}, semantic knowledge \cite{theiner2022interpretable, haas2024pigeon}, or a combination of approaches (e.g., hierarchy and semantic knowledge \cite{haas2024pigeon}). 

In addition to partitioning, many classification-based methods explored how to improve performance by incorporating scene knowledge or semantics. Individual Scene Networks (ISNs) were used to diversify learned features across indoor, urban, and natural scenes \cite{muller2018geolocation}. GeoDecoder further diversified features across both geographic levels and 16 visual scene categories \cite{clark2023we}. IM2City \cite{wu2022im2city} introduced CLIP and visual-language grounding, by training with images paired with their city names. In G$^3$, image embeddings and clue embeddings from human-written GeoGuessr guidebooks were combined to predict the country of the query image \cite{luo2022g3}. TransLocator \cite{pramanick2022world} performs repeated fusion across parallel semantic and RGB backbones.

In GeoSURGE, we use semantic segmentations as latent cross-attention \cite{guo2025deepseek} queries  to guide RGB feature aggregation. This treats semantics not as a learned second representation, but as a structural signal for correspondence. While we use many concepts from classification-based methods such as geocells and hierarchy, we treat visual geo-localization like a retrieval problem wherein we match visual and geographic representations.

\textbf{Retrieval-based methods} typically compare a query image to a large database of geotagged images to infer the location of the query image \cite{hays2008im2gps, hays2015large}. The seminal work of IM2GPS first investigated planet-scale geo-localization via retrieval by computing similarities between hand-crafted features of the query image and a geotagged image database over 6 million images \cite{hays2008im2gps}. Recently Img2Loc \cite{zhou2024img2loc} leverages LVLMs, such as GPT-4V, extending the retrieval-based approach with retrieval augmented generation (RAG). After retrieving the most and least similar reference images to the query image, Img2Loc prompts an LVLM to geotag the query image. In this way, Img2Loc benefits from the Internet-scale multimodal corpus used to train LVLMs to infer image coordinates from visual cues.

GeoCLIP \cite{vivanco2024geoclip} reformulated the retrieval-based approach as comparing the query image to GPS coordinates using CLIP-style contrastive learning, effectively opting for a reference database of GPS coordinates instead of geotagged images. Their design choice allows the network to accumulate visual information from multiple scenes associated with the same coordinates, in essence learning a representation of the location. Similar in spirit to Img2Loc, G3 \cite{jia2024g3} extends GeoCLIP's retrieval-based approach with RAG using LVLMs, such as GPT-4V. Recognizing that LLMs struggle for specialized tasks, such as geolocalization \cite{campos2026gaea}, GeoReasoner \cite{li2024georeasoner} and GAEA \cite{campos2026gaea} have devised new approaches and training data to better leverage the powerful reasoning capabilities of LLMs for geo-localization.

In GeoSURGE, we also adopt CLIP-style contrastive learning but differ in how geographic representations are defined. Instead of embedding GPS coordinates as in GeoCLIP, we learn feature embeddings for geographic cells and train them jointly with image embeddings under a contrastive objective. This design avoids the limitations of low-dimensional GPS representations, which require specialized components such as Random Fourier Features in GeoCLIP. Empirically, this design choice leads to improved geo-localization performance compared to GeoCLIP while using the same backbone architectures and training data.

\textbf{Hybrid Methods} aim to merge retrieval and classification to improve performance. [L]kNN use features derived from networks trained with a classification learning objective in their retrieval-based inference system \cite{vo2017revisiting}. In \cite{kordopatis2021leveraging} and \cite{astruc2024openstreetview}, a retrieval-within-geocell scheme is used to refine the predicted location of the query image inside the S2 geocell, using features trained via classification. PIGEOTTO \cite{haas2024pigeon} refines the predicted location of the query image through a hierarchical retrieval mechanism that examines both top-K geocells and clusters within these geocells. GeoSURGE can also be considered a hybrid as it is trained to match scene features with geographic features (i.e., retrieval) that represent geocells (i.e., classification).

%% file: sec/3_approach.tex
\section{Approach}
\label{sec:approach}

GeoSURGE is trained to geolocate images through contrastive learning by aligning image features with geographic features that represent regions of Earth as in Figure~\ref{fig:overview}. The visual representation of an image is derived from both the RGB image and its semantic segmentation map, which are assumed inputs. For training, we also assume access to a dataset of geotagged images. The geographic representation is a learned hierarchy of embeddings that correspond to regions of Earth. Geo-localization is implemented by outputting the location of the geographic representation that contains a query image, which is unknown. Therefore, GeoSURGE learns a function to maximize the similarity between query image features and the geographic features corresponding to the region of Earth that contains that image. Below we detail the design of these representations, our training procedure, and how we use these two representations for inference in a hierarchical manner.

\subsection{Geographic Representation}
\label{sec:approach_geocell}

GeoSURGE models geography using a hierarchical, distributed representation. The Earth's surface is first partitioned into geographic cells (geocells), similar to classification-based geo-localization methods. However, instead of treating geocells as discrete class labels, GeoSURGE learns a feature embedding for each geocell using all training samples assigned to that region. Collectively, these embeddings form a distributed representation of geographic space. Inspired by hierarchical geo-localization approaches \cite{vo2017revisiting,muller2018geolocation}, GeoSURGE constructs multiple such partitions at different spatial resolutions. The resulting geocells therefore form a hierarchy of partitions, which we represent as a hierarchy of learned embedding spaces.

In GeoSURGE we use Google’s S2 Geometry Library to divide Earth's surface into geocells \cite{muller2018geolocation}. The process begins by projecting the Earth onto the six faces of a cube, resulting in six initial S2 geocells containing all training samples from a dataset. To balance the number of images within each geocell, any geocell containing more than $\tau_{\textrm{max}}$ samples is recursively subdivided. Geocells with fewer than $\tau_{\textrm{min}}$ samples are excluded to ensure that each geocell has a sufficient number of samples. This recursive splitting continues until all geocells contain a number of samples between $\tau_{\textrm{min}}$ and $\tau_{\textrm{max}}$. This way ultimately produces a partition~$\rho$ of Earth's surface where each geocell contains a balanced number of training samples, $\rho(\tau_{\textrm{min}},\tau_{\textrm{max}})$. GeoSURGE then explicitly models each geocell as a unique vector that represents all training samples within. Together, these vectors form the embedding space, $\mathcal{E}$, representing the partition of Earth's surface $\rho(\tau_{\textrm{min}},\tau_{\textrm{max}})$ that is learned during training. Note, GeoSURGE is agnostic to different partitioning schemes and uses S2 partitioning \cite{muller2018geolocation} because our contributions are orthogonal to the partitioning method.

GeoSURGE constructs a hierarchical partition of Earth by recursively applying the above partitioning process at multiple granularities, ultimately representing each geocell partition as a distinct embedding. We treat the number of partitions, the dimensionality of the embedding space representing a partition, and $(\tau_{\textrm{min}},\tau_{\textrm{max}})$ for each partition as hyperparameters. During repeated partitioning, the value of $\tau_{\textrm{max}}$ is varied while the same value of $\tau_{\textrm{min}}$ is applied across all hierarchy partitions. This approach ensures that each geocell in the finest partition can be uniquely linked to its coarser parent partitions. Therefore, at inference time a hierarchical prediction is computed as the product of similarities between features of a query image and all containing geocells of the partition hierarchy (i.e., embeddings of corresponding geocells). However, during training, the embedding representing each partition is learned separately from other partitions to promote diversity among geographical features at different scales. We describe more details on training the hierarchy of geographic embeddings and how to use the learned hierarchy of geographic embeddings for inference in later sections.

\begin{figure}[t]
  \fontsize{10}{10}\selectfont
  \centering
   \includegraphics[width=0.8\linewidth]{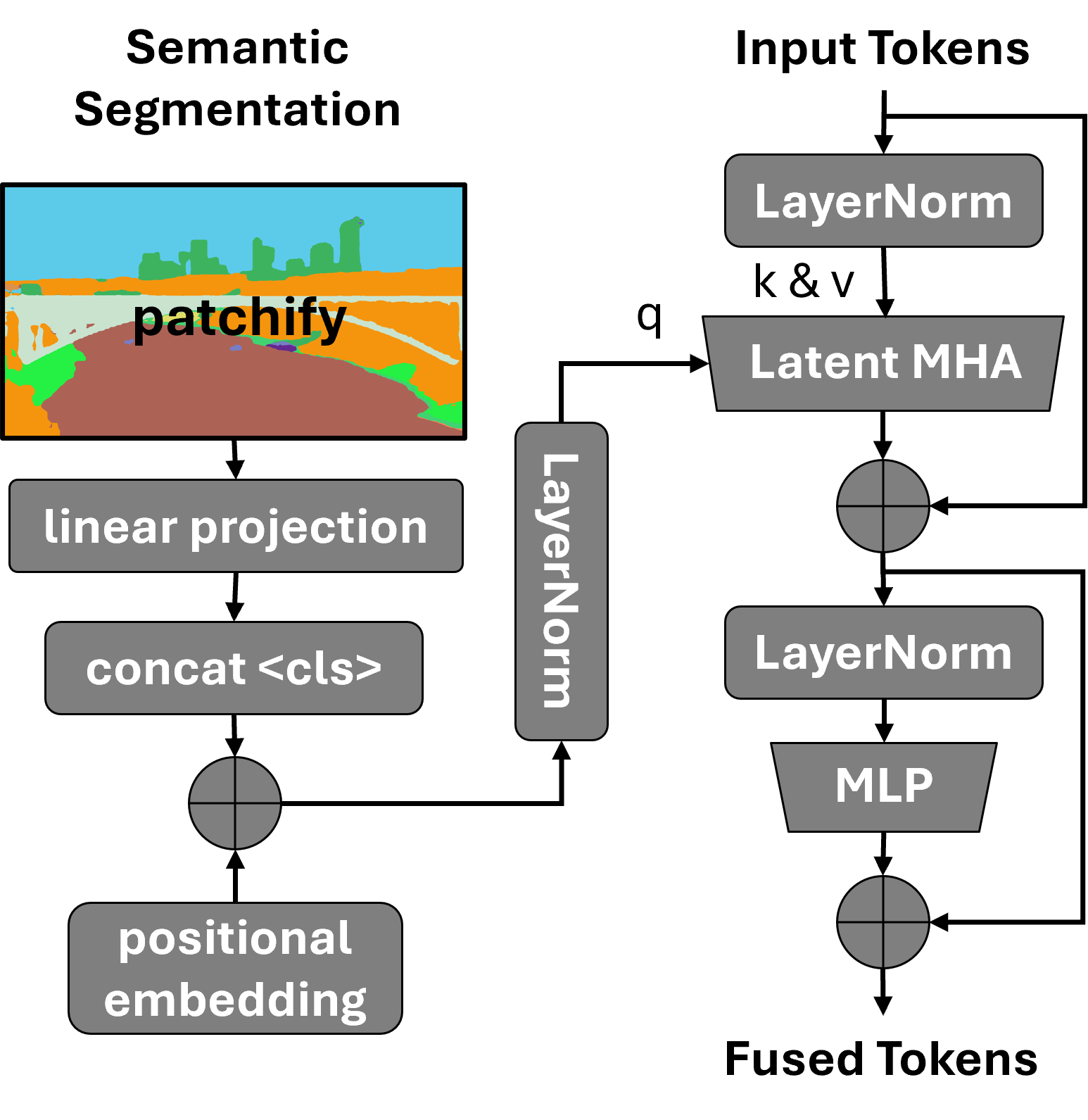}
   \vspace{-0.5cm}
   \caption{Diagram of GeoSURGE's semantic fusion blocks.}
   \vspace{-0.5cm}
   \label{fig:fusion}
\end{figure}

\subsection{Visual Representation}
\label{sec:approach_visual}

GeoSURGE extracts visual features from both appearance (RGB image) and semantics (segmentation map) to match with learned features from the geographic representation. We use CLIP \cite{radford2021learning} to extract features from the RGB image as in \cite{vivanco2024geoclip,haas2024pigeon,jia2024g3}. We then uses OneFormer \cite{jain2023oneformer} to get a semantic segmentation map corresponding to this RGB image. This semantic segmentation map is fused with the CLIP-extracted features through latent cross-attention within a semantic fusion module, as shown in Figure~\ref{fig:fusion}. The fused visual representation output from this semantic fusion module is then matched to the learned geographic features for predicting the image's location.

The semantic fusion module combines features extracted from the RGB image and its semantic segmentation map into a single visual representation. GeoSURGE first uses the CLIP vision transformer (i.e., without projection and output normalization) to extract RGB appearance features as a sequence of tokens representing all patches of the RGB image plus a CLS token. All but the last CLIP Vision Transformer encoder blocks are kept frozen during training. The semantic fusion module enhances these RGB tokens, by fusing them with a semantic segmentation map produced from the RGB image. We select OneFormer \cite{jain2023oneformer} for semantic segmentations to balance between performance and compute time as we need to preprocess (segment) hundreds of thousands of images across datasets. The semantic segmentation map has the same width and height as the RGB image and semantic classes are from ADE20K \cite{zhou2019semantic}. 

Shown in the left of Figure~\ref{fig:fusion}, the semantic fusion module extracts tokens from the semantic segmentation map, by linearly projecting all patches from the segmentation map, concatenating a CLS token, and adding positional embeddings. We fuse the semantic tokens with the RGB tokens using latent multi-headed attention to promote memory efficiency \cite{guo2025deepseek}. RGB image tokens are used as the keys and values, while semantic tokens are queries. The output of this attention module is then passed through an MLP, producing fused tokens for every patch of the image. Additionally, the semantic fusion module includes residual connections \cite{he2016deep} and layer normalizations common in transformer architectures, as shown in Figure~\ref{fig:fusion}. Although Figure~\ref{fig:fusion} shows only a single fusion block, the semantic fusion module serially repeats these to promote hierarchies in fused features. As final steps, we extract the CLS of the fused tokens, then perform layer normalization and linear projection of the fused CLS token to produce the final visual representation.

\subsection{Training}
\label{sec:approach_learning}

GeoSURGE aims to align visual features from the query image with the learned geographic features associated with that image's specific location. The geographic features are embeddings representing a partition hierarchy of Earth's surface. The visual features are extracted from the input RGB image and its semantic segmentation map. Our training process uses contrastive learning techniques \cite{radford2021learning} to maximize the similarity of correct pairs of visual and geographic features from the training dataset, while minimizing the similarity of incorrect pairs.

Specifically, GeoSURGE is trained to select the correct geographic location for a query image from a batch that also contains incorrect locations, by computing the similarity between the features of each location and the query image. Our training batches are constructed from sampling a dataset $\mathcal{D}$ of geotagged images $(x, y)$. Given a training sample $x$, we extract the fused CLS token as a feature vector $\textbf{v}$, summarizing the RGB and the semantic information of the sample. We then use the sample's true location $y$ to extract a second feature vector $\textbf{g}$ from our geographic representation. This feature vector is a parameter learned through the training objective. We construct $\textbf{g}$, by indexing the partition $\rho(\tau_{\textrm{min}},\tau_{\textrm{max}})$ for the geocell containing $y$. The corresponding normalized vector of the geocell within the embedding $\mathcal{E}$ representing $\rho(\tau_{\textrm{min}},\tau_{\textrm{max}})$ is $\textbf{g}$. 

Given a batch of correct pairs of visual and geographic features $(\textbf{v},\textbf{g})$ of size $B$, we learn a function that maximizes the cosine similarity of the $B$ correct pairs and minimizes the cosine similarity of the $B^2-B$ incorrect pairs. Our learning objective is the InfoNCE loss function \cite{oord2018representation} shown in Equation~\ref{eq:loss} for a sample $i$ in a batch. Following the above procedure, we extract feature vectors for each partition within our hierarchical geographic representation. The complete learning objective is the sum of losses for the full partition hierarchy.
\begin{equation}
    \label{eq:loss}
    \mathcal{L}_i=-\textrm{log}\frac{\textrm{exp}(\textbf{v}^{\intercal}_i\textbf{g}_i / \tau)}{\textrm{exp}(\textbf{v}^{\intercal}_i\textbf{g}_i / \tau)+\sum_{j\neq i}\textrm{exp}(\textbf{v}^{\intercal}_i\textbf{g}_j / \tau)}
\end{equation}

\subsection{Hierarchical Inference}
\label{sec:approach_inference}

During inference (Figure~\ref{fig:overview}), GeoSURGE matches the query image's visual features with all features of the learned geographic representation. GeoSURGE uses a hierarchical inference process to produce a single prediction that integrates the complete partition hierarchy of Earth’s surface. In other words, the location of the query image is predicted via this hierarchical inference process as the location in the partition hierarchy of Earth's surface with the highest similarity to the query image's visual features.

Specifically, given a query image $x$ with unknown location, GeoSURGE extracts the fused CLS token of its visual representation as a feature vector $\textbf{v}$. The extraction process is described in previous sub-sections. GeoSURGE then computes the cosine similarity between $\textbf{v}$ and the vectors of each embedding $\mathcal{E}$, which represent the partitions $\rho(\tau_{\textrm{min}},\tau_{\textrm{max}})$ dividing Earth's surface in a hierarchical manner. These similarities are then normalized using softmax to get probabilities. For each geocell $r$ in the finest partition, we integrate the probabilities for each parent geocell $r'$ that contains $r$ by computing the product of their probabilities. This way integrates similarities across all hierarchy levels, realizing hierarchical inference.

%% file: sec/4_experiments.tex
\section{Experiments}
\label{sec:eval}

We perform experiments on publicly available benchmark datasets to have a fair comparison with existing works. We provide additional ablation studies to gauge the contribution of our design choices to overall performance. We observed that GeoSURGE provides state-of-the-art results in 22 / 25 metrics measured across five benchmark datasets. Ablation studies indicate that these improvements are mostly due to our geographic and visual representations.

\textbf{Benchmark Evaluation Protocol} follows the precedent established in previous works \cite{clark2023we,jia2024g3,muller2018geolocation,pramanick2022world,vivanco2024geoclip,zhou2024img2loc} to maintain fair comparisons. We train GeoSURGE on the MediaEval Placing Tasks 2016 (MP-16) dataset containing more than 4 million images, holding 1\% for validation and the remainder for training. We tested GeoSURGE on five benchmark datasets: IM2GPS \cite{hays2008im2gps}, IM2GPS3k \cite{vo2017revisiting}, YFCC4k \cite{vo2017revisiting} and YFCC26k \cite{muller2018geolocation}, and GWS15k \cite{clark2023we}. Note, while the GWS15k image dataset has not been publicly released, we use the same set of panorama IDs from the authors that uniquely identify the Google Street View image panoramas from where the original GWS15k was constructed. In this way, our results on GWS15k can be directly compared with \cite{clark2023we,vivanco2024geoclip,campos2026gaea} and avoid differences in results reported from \cite{haas2024pigeon} that may be introduced due to random sampling (result indicated in {\color{gray}gray}). Same as \cite{clark2023we}, we average the prediction of the Ten Crop method to provide a single prediction for the entire image. As in prior work, we report results using a threshold metric. We computed the great circle distance (GCD) from each predicted location to the ground truth location using the Haversine distance. After computing the GCDs for all predicted coordinates with respect to the ground truth, we compute the percentage of predictions within five error thresholds: 1km, 25km, 200km, 750km, 2500km. These thresholds correspond to 5 levels of localization: street, city, region, country, continent.

\textbf{Implementation details} are summarized as follows to support recreation of results. We use Clip-ViT-Large-Patch14-336 as our visual backbone. The semantic fusion module extracts 128-dimensional tokens from the semantic segmentation maps by linearly projecting each 14 by 14 patch and uses 3 repeated fusion blocks with query dimensionalities of 128. Remaining dimensionalities of the semantic fusion module match the hidden dimensionality of visual backbone, 1024, to perform latent cross-attention. The latent dimension of 64 was used for cross-attention. We used 7 partitioning levels to divide Earth where $\tau_{min}$ is 50 and $\tau_{max}$ is 25000, 10000, 5000, 2000, 1000, 750, 500 from coarsest to finest partitioning. We initialized the learnable temperature parameters in Equation~\ref{eq:loss} to 0.07 for each partition level with geographic embeddings having a dimensionality of 768 to match the visual backbone's projection dimensionality. We trained using the AdamW optimizer with initial learning rate 0.0001, weight decay 0.0001, and effective batch size of 1024. We used a step learning rate decay schedule with a gamma of 0.5 at each epoch. We used early stopping to finish training when the performance on the validation dataset did not improve for 4 epochs. GeoSURGE was trained in 21 hours on 8 NVIDIA A6000 GPUs.

\subsection{Experimental Results}
\label{sec:eval_results}

Tables \ref{tbl:im2gps} through \ref{tbl:gws15k} summarize our performance on five benchmark datasets. We bold the best result and underline the second best for each distance threshold in a dataset. Compared prior global visual geo-localization works include {[}L{]}kNN \cite{vo2017revisiting}, PlaNet \cite{weyand2016planet}, CPlaNet \cite{seo2018cplanet}, ISNs \cite{muller2018geolocation}, Translocator \cite{pramanick2022world}, GeoDecoder \cite{clark2023we}, GeoCLIP \cite{vivanco2024geoclip}, and PIGEOTTO   \cite{haas2024pigeon}, Img2Loc \cite{zhou2024img2loc}, G3 \cite{jia2024g3}, RFM$_{\textrm{\tiny 10M}}\mathcal{S}_2$ \cite{dufour2025around}, GeoReasoner (GR/Qwen-VL) \cite{li2024georeasoner}, and GAEA (GA/Qwen2.5-VL) \cite{campos2026gaea}. If a method was published before a benchmark dataset was released, we do not include it in the table correspondent to that benchmark dataset. In \textit{italics} we provide results we generated for prior work when code and pretrained weights were available in an effort to provide more complete benchmarks. We place "-" in the table for remaining unavailable results.

\textbf{Quantitative results} show GeoSURGE achieves state-of-the-art in 22 / 25 metrics across the five distance thresholds and five benchmark datasets. Excluding methods that use LVLMs, GeoSURGE performs better in all 25 metrics measured across five benchmark datasets. The performance gains over SOTA GeoCLIP method highlight the importance of GeoSURGE's geographic representation design as GeoCLIP uses the same visual backbone with a GPS-based geographic representation. Our approach learns hierarchical geographic representations that capture multi-scale structure along with semantic and relational context between regions and their neighbors. We further probe the performance benefits attributable to GeoSURGE's geographic representation in ablation studies.

When considering recent LVLM-based methods, GeoSURGE outperforms both Img2Loc and G3 in 7 / 10 metrics across five distance thresholds and two benchmark datasets reported in \cite{zhou2024img2loc, jia2024g3}. Specifically, Img2Loc or G3 in some cases perform better than GeoSURGE with smaller distance thresholds (street and city-level) for evaluation. As these large-scale models are exposed to a massive corpora of multimodal data, gains at finer-grain levels may be attributed to latent memorization of common visual cues or pattern recognition within specific locales. For instance, the GPT-4V based-models may recognize specific landmark features such as signage, architectural design, or vegetation patterns from Internet-scale tourist photos to accurately pinpoint street or city-level position. These results indicate that explicit geographic modeling remains crucial to global visual geolocation, even as LVLMs continue to advance.



\begin{figure}[t]
  \fontsize{10}{10}\selectfont
  \centering
   \includegraphics[width=0.85\linewidth]{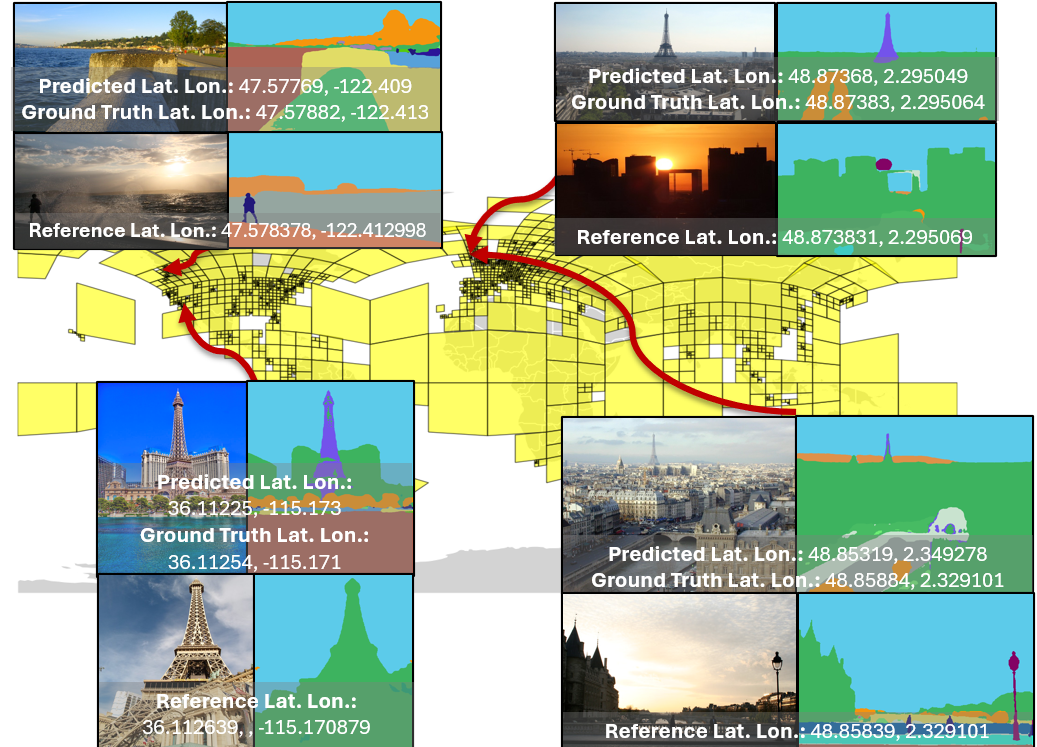}
   \caption{Examples of successful predictions (GeoSURGE). Best viewed when zoomed.}
   \label{fig:qual_correct}
\end{figure}

\textbf{Qualitative results} of sample success cases are shown in Figure~\ref{fig:qual_correct}. We show the predicted GPS, ground truth GPS, as well as the closest reference image to the predicted location. We show that GeoSURGE is robust in location prediction, by showcasing an example with the Eiffel Tower (the bottom-left example in Figure~\ref{fig:qual_correct}): the original in Paris and replica in Las Vegas. Distinctive features derived from the semantic information assists the model in differentiating these two locations, such as the water and large building located close to the replica that are absent at the original location. Dynamic entities, such as the person in the top-left reference image and two people in the bottom-right reference image in Figure~\ref{fig:qual_correct}, are also accurately identified in semantic segmentation maps. This way implicitly avoids the utilization of unreliable imaged regions to geo-localization, via our semantic fusion module.

\begin{table}[t] 
    \fontsize{9}{9}\selectfont
    \setlength{\tabcolsep}{1mm}
    \centering
    \caption{IM2GPS GCD accuracy; higher is better.}
    \vspace{-0.3cm}
    \begin{tabular}{lccccc}
        \toprule
         & Street & City  & Region & Country & Continent \\
        Method & 1 km   & 25 km & 200 km & 750 km  & 2500 km   \\
        \midrule
        {[}L{]}kNN                                 & 14.4   & 33.3  & 47.7   & 61.6    & 73.4      \\
        PlaNet                                      & 8.4    & 24.5  & 37.6   & 53.6    & 71.3      \\
        CPlaNet                                    & 16.5   & 37.1  & 46.4   & 62.0    & 78.5      \\
        ISNs                                      & 16.9   & 43.0  & 51.9   & 66.7    & 80.2      \\
        Translocator                                & 19.9   & 48.1  & 64.6   & 75.6    & 86.7      \\
        GeoDecoder                                 & \underline{22.1}   & \underline{50.2}  & \underline{69.0}   & 80.0    & 89.1      \\
        GeoCLIP                                      &  \textit{16.5}      & \textit{40.9}     &  \textit{54.9}      &  \textit{76.8}       &  \textit{88.6}         \\
        PIGEOTTO                                     & 11.8   & 38.8  & 63.7   & \underline{80.5}    & \underline{91.1}      \\
        RFM$_{\textrm{\tiny 10M}}\mathcal{S}_2$ & \textit{8.0}      & \textit{44.7}     & \textit{64.1}      & \textit{78.5}       & \textit{88.2}         \\
        \midrule
        Img2Loc/GPT-4V & -      & -     & -      & -       & -         \\
        G3/GPT-4V                                         & -
        & -     & -      & -       & -         \\
        \small GR/Qwen-VL & \textit{8.9} & \textit{41.1} & \textit{56.2} & \textit{72.3} & \textit{88.4}         \\
        \small GA/Qwen2.5-VL & -      & 43.0 & 57.4 & 77.2 & 89.5 \\
    
        \midrule
        GeoSURGE (Ours)                                    & \textbf{27.0} & \textbf{54.4} & \textbf{70.0}   & \textbf{84.4} & \textbf{93.2}     \\
         \bottomrule

    \end{tabular}
    
    \label{tbl:im2gps}
\end{table}

\begin{table}[t] 
    \fontsize{9}{9}\selectfont
    \setlength{\tabcolsep}{1mm}
    \centering
    \caption{IM2GPS3k GCD accuracy; higher is better.}
    \vspace{-0.3cm}
    \begin{tabular}{lccccc}
        \toprule
         & Street & City  & Region & Country & Continent \\
        Method & 1 km   & 25 km & 200 km & 750 km  & 2500 km   \\
        \midrule
        {[}L{]}kNN               & 7.2    & 19.4  & 26.9   & 38.9    & 55.9      \\
        PlaNet                & 8.5    & 24.8  & 34.3   & 48.4    & 64.6      \\
        CPlaNet                 & 10.2   & 26.5  & 34.6   & 48.6    & 64.6      \\
        ISNs                   & 10.5   & 28.0  & 36.6   & 49.7    & 66.0      \\
        Translocator            & 11.8   & 31.1  & 46.7   & 58.9    & 80.1      \\
        GeoDecoder           & 12.8   & 33.5  & 45.9   & 61.0    & 76.1      \\
        GeoCLIP                 & 14.1   & 34.5  & 50.7   & 69.8    & 83.8      \\
        PIGEOTTO               & 10.9   & 35.8  & 52.4   & 70.7    & 84.4      \\
        RFM$_{\textrm{\tiny 10M}}\mathcal{S}_2$ & \textit{6.0}      & \textit{36.8}     & \textit{51.0}      & \textit{66.9}       & \textit{81.6}         \\
        \midrule
        Img2Loc/GPT-4V                 & \underline{17.1}   & \textbf{45.1}  & \underline{57.9}   & 72.9    & 84.7      \\

        G3/GPT-4V                & 16.6   & 40.9  & 55.6   & 71.2    & 84.7      \\
        \small GR/Qwen-VL & \textit{8.8} & \textit{33.4} & \textit{44.6} & \textit{61.3} & \textit{78.7} \\
        \small GA/Qwen2.5-VL & - & 36.9 & 56.0 & \underline{73.2} & \underline{86.7} \\
        \midrule
        GeoSURGE (Ours)                    & \textbf{17.2}  & \underline{42.5}  & \textbf{58.1}   & \textbf{74.6}    & \textbf{87.6}     \\
        
        \bottomrule

    \end{tabular}
    \label{tbl:im2gps3k}
\end{table}

\begin{table}[t] 
    \fontsize{9}{9}\selectfont
    \setlength{\tabcolsep}{1mm}
    \centering
    \caption{YFCC4k GCD accuracy; higher is better.}
    \vspace{-0.3cm}
    \begin{tabular}{lccccc}
        \toprule
         & Street & City  & Region & Country & Continent \\
        Method                & 1 km   & 25 km & 200 km & 750 km  & 2500 km   \\
        \midrule
        {[}L{]}kNN                                 & 2.3           & 5.7           & 11.0          & 23.5          & 42.0          \\
        PlaNet                                    & 5.6           & 14.3          & 22.2          & 36.4          & 55.8          \\
        CPlaNet                                    & 7.9           & 14.8          & 21.9          & 36.4          & 55.5          \\
        ISNs                                       & 6.7           & 16.5          & 24.2          & 37.5          & 54.9          \\
        Translocator                                & 8.4           & 18.6          & 27.0          & 41.1          & 60.4          \\
        GeoDecoder                                   & 10.4          & 24.4          & 33.9          & 50.0          & 68.7          \\
        GeoCLIP                                     &  \textit{10.1} &  \textit{19.9}             &  \textit{34.2}             &  \textit{56.1}             &  \textit{75.3}             \\
        PIGEOTTO                                   & 9.5           & 22.5          & 38.8          & 60.7          & 76.9          \\
        RFM$_{\textrm{\tiny 10M}}\mathcal{S}_2$ & \textit{6.6}      & 33.5     & 45.3      & 61.1       & 77.7         \\
        \midrule
        Img2Loc/GPT-4V                           & 14.1          & 29.6          & 41.4          & 59.3          & 76.9          \\
        G3/GPT-4V                 & \textbf{24.0}   & \textbf{35.9}  & \underline{47.0}   & \underline{64.3}    & \underline{78.1}      \\
        \small GR/Qwen-VL & \textit{2.0} & \textit{10.2} & \textit{18.0} & \textit{37.1} & \textit{59.8} \\
        \small GA/Qwen2.5-VL & - & - & - & - & - \\
\midrule
        GeoSURGE (Ours)                                        &  \underline{19.9} & \underline{33.6} & \textbf{48.7} & \textbf{67.4} & \textbf{82.0}     \\
        \bottomrule
    \end{tabular}
    \label{tbl:yfcc4k}
\end{table}

\begin{table}[t] 
    \fontsize{9}{9}\selectfont
    \setlength{\tabcolsep}{1mm}
    \centering
    \caption{YFCC26k GCD accuracy; higher is better.}
    \vspace{-0.3cm}
    \begin{tabular}{lccccc}
        \toprule
         & Street & City  & Region & Country & Continent \\
        Method & 1 km   & 25 km & 200 km & 750 km  & 2500 km   \\
        \midrule
        PlaNet                                     & 4.4           & 11.0          & 16.9          & 28.5          & 47.7          \\
        ISNs                                        & 5.3           & 12.3          & 19.0          & 31.9          & 50.7          \\
        Translocator                                & 7.2           & 17.8          & 28.0          & 41.3          & 60.6          \\
        GeoDecoder                                  & 10.1          & 23.9          & 34.1          & 49.6          & 69.0          \\
        GeoCLIP                                    & \underline{11.6}          & 22.2          & 36.7          & 57.5          & 76.0          \\
        PIGEOTTO                                   & 10.1          & 24.6          & \underline{41.3}          & \underline{62.6}          & \underline{78.7}          \\
        RFM$_{\textrm{\tiny 10M}}\mathcal{S}_2$ & \textit{5.3}      & \textit{\underline{29.0}}     & \textit{40.9}      & \textit{57.8}       & \textit{75.8}         \\
         \midrule
        Img2Loc/GPT-4V                             & -             & -             & -             & -             & -             \\
        G3/GPT-4V                            & -             & -             & -             & -             & -             \\
        \small GR/Qwen-VL & \textit{4.0} & \textit{17.4} & \textit{28.9} & \textit{48.1} & \textit{67.8} \\
        \small GA/Qwen2.5-VL & - & - & - & - & - \\
\midrule
        GeoSURGE (Ours) & \textbf{17.8} & \textbf{31.5} & \textbf{45.1} & \textbf{64.3} & \textbf{79.3}     \\
        \bottomrule
    \end{tabular}
    \label{tbl:yfcc26k}
\end{table}

\begin{table}[t!] 
    \fontsize{9}{9}\selectfont
    \setlength{\tabcolsep}{1mm}
    
    \caption{\centering GWS15k GCD accuracy; higher is better.}
    \vspace{-0.3cm}
    \begin{tabular}{lccccc}
        \toprule
         & Street & City  & Region & Country & Continent \\
        Method & 1 km   & 25 km & 200 km & 750 km  & 2500 km   \\
        \midrule
        GeoDecoder                                & \underline{0.7} & 1.5          & 8.7           & 26.9          & 50.5          \\
        GeoCLIP                                     & 0.6          & 3.1          & \underline{16.9} & \underline{45.7} & \underline{74.1}          \\
        PIGEOTTO$\dagger$ & {\color[HTML]{9B9B9B}0.1} & {\color[HTML]{9B9B9B}8.7} & {\color[HTML]{9B9B9B}30.1} & {\color[HTML]{9B9B9B}64.0} & {\color[HTML]{9B9B9B}84.7} \\
       RFM$_{\textrm{\tiny 10M}}\mathcal{S}_2$ & \textit{0.1}  & \textit{2.6}     & \textit{15.0}      & \textit{42.2}       & \textit{70.0}         \\
        \midrule
        Img2Loc/GPT-4V & - & - & - & - & - \\
        G3/GPT-4V & - & - & - & - & - \\
        \small GR/Qwen-VL & \textit{0.3} & \textit{3.0} & \textit{16.1} & \textit{43.6} & \textit{71.7} \\
        \small GA/Qwen2.5-VL & - & \underline{3.7} & 16.7 & 43.3 & 73.5 \\
    \midrule
        GeoSURGE (Ours)                                       & \textbf{1.0}           & \textbf{4.6} & \textbf{21.9} & \textbf{54.7} & \textbf{80.8} \\
        \bottomrule
    \end{tabular}
    {\footnotesize \rightline{$\dagger$See Benchmark Evaluation Protocol in Section~\ref{sec:eval}}}
    \label{tbl:gws15k}
\end{table}

\begin{figure}[t]
  \fontsize{10}{10}\selectfont
  \centering
   \includegraphics[width=0.85\linewidth]{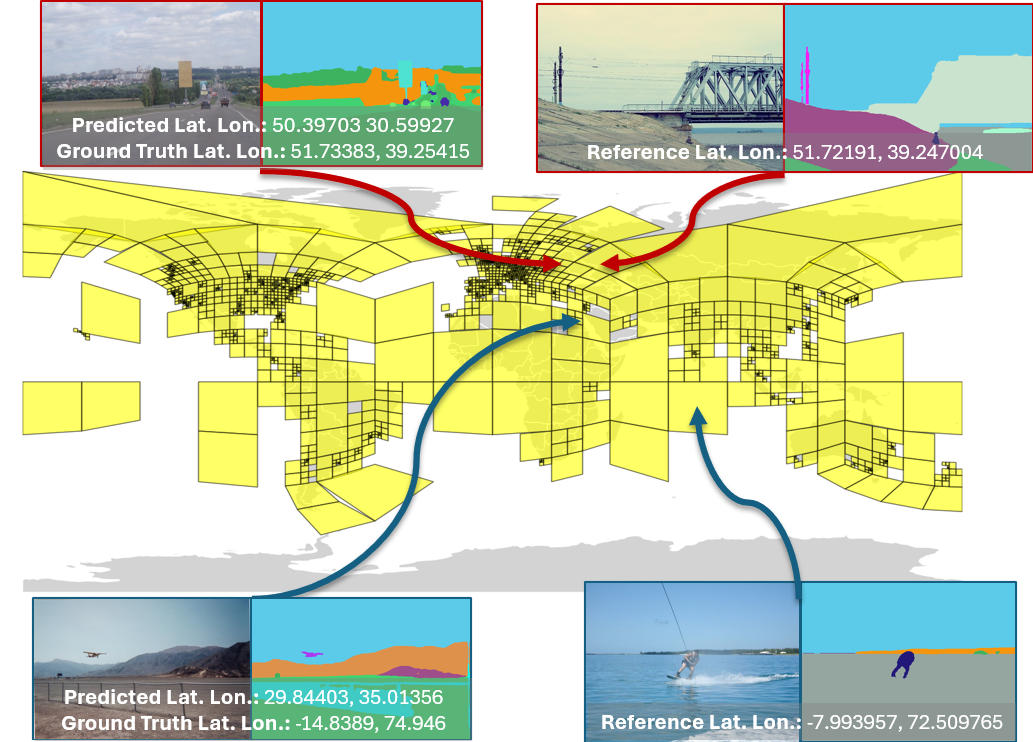}
   \caption{Examples of unsuccessful predictions (GeoSURGE). Best viewed when zoomed.}
   \label{fig:qual_incorrect}
\end{figure}

We also report some failure cases from GeoSURGE in Figure \ref{fig:qual_incorrect}. Here, we show predicted reference-image pairs, denoted by color. For the pair denoted in red, while within country-level distance between the predicted location and ground truth, visual ambiguity with the RGB image and semantic image may lead to confusion in the prediction. For the pair denoted in blue, the large disparity in predicted latitude and longitude versus ground truth can be attributed to an insufficient number of training images within the ground truth cell. The closest reference image to the ground truth is approximately 806 kilometers away.

\subsection{Ablation Studies}
\label{sec:eval_ablations}

We conducted ablation studies using the YFCC26k and GWS15k datasets with different settings to our design choices in GeoSURGE. The YFCC26k datatset was used for ablations because YFCC26k, IM2GPS, IM2GPS3k, and YFCC4k are sourced from Yahoo Flikr Creative Commons 100 Million (YFCC100M), and YFCC26k contains the most testing images (i.e., 25,600). In this way, YFCC26k is most representative of this benchmark group. Ablations are also reported for the GWS15k benchmark dataset because its testing image distribution differs from the YFCC dataset, being sourced from Google Streetview and designed to be more uniformly distributed across Earth \cite{clark2023we}. Tables~\ref{tbl:ablation_hier} through \ref{tbl:ablation_geo2} show the impact of GeoSURGE's design choices to the overall performance for these benchmarks. Ablations for remaining datasets are in supplementary materials.

\textbf{Hierarchy depth ablations} are shown in Table~\ref{tbl:ablation_hier}, where we vary the depth of the geographic hierarchy GeoSURGE uses during training and inference. All other variables are set to defaults and held constant across the GeoSURGE ablations. We see that deeper hierarchies improve performance, as observed in prior works. This stems from having smaller geocells with more precise locations as well as more diversity among geographical features at different scales. Consequently, these trends are most pronounced for the street and city level metrics.

\textbf{Semantic fusion ablations} are shown in Table~\ref{tbl:ablation_sem} where we provide the GCD accuracy when both activating the semantic fusion module with varying number of fusion blocks (i.e., 1 to 3 fusion blocks) and disabling semantic fusion altogether to isolate its performance contribution (i.e., None fusion blocks). All other variables are set to defaults and held constant across the GeoSURGE ablations. Across all five benchmarks we observe significant improvements between 0 versus 3 fusion blocks (e.g., average 36.5\% relative improvement for 1km GCD accuracy). Generally, accuracy tends to increase with number of fusion blocks. We selected 3 fusion blocks in GeoSURGE to balance performance across benchmarks.

\textbf{Geographic representation ablations} are shown in Tables~\ref{tbl:ablation_geo} and ~\ref{tbl:ablation_geo2}, where we toggle whether GeoSURGE is trained with a retrieval learning objective that uses our geographic embeddings or a classification learning objective that does not. In Table~\ref{tbl:ablation_geo}, the geographic embeddings are toggled with the full hierarchy of GeoSURGE, while in Table~\ref{tbl:ablation_geo2} only the finest partition level of the hierarchy is used to emulate a flat representation. Note, this setting differs from the last rows in Table~\ref{tbl:ablation_hier}, which used the middle level of the hierarchy for fair comparison. All other variables are set to defaults and held constant across these two sets of ablations.

\begin{table}[t] 
    \fontsize{9}{9}\selectfont
    \setlength{\tabcolsep}{1mm}
    \centering
    \caption{Hierarchy Depth Ablations}
    \vspace{-0.3cm}
    \begin{tabular}{cccccc}
        \toprule
        \multicolumn{6}{c}{YFCC26k GCD accuracy; higher is better} \\
        \midrule
        Hierarchy & Street & City  & Region & Country & Continent \\
        Levels & 1 km   & 25 km & 200 km & 750 km  & 2500 km   \\
        \midrule
        7 & 17.8 & 31.5 & 45.1 & 64.3 & 79.3 \\
        5 & 11.1 & 30.0 & 44.4 & 62.1 & 77.5 \\
        3 & 10.4 & 28.9 & 43.7 & 61.7 & 77.4 \\
        1 &  8.9 & 27.5 & 42.9 & 61.5 & 77.2 \\
        \midrule
        \multicolumn{6}{c}{GWS15k GCD accuracy; higher is better} \\
        \midrule
        Hierarchy & Street & City  & Region & Country & Continent \\
        Levels & 1 km   & 25 km & 200 km & 750 km  & 2500 km   \\
        \midrule
        7 & 1.0 & 4.6 & 21.9 & 54.7 & 80.8 \\
        5 & 0.4 & 3.5 & 21.1 & 53.6 & 79.7 \\
        3 & 0.5 & 3.9 & 22.1 & 54.7 & 80.1 \\
        1 & 0.1 & 3.1 & 20.8 & 52.5 & 79.3 \\
        \bottomrule
    \end{tabular}
    \label{tbl:ablation_hier}
\end{table}

\begin{table}[t] 
    \fontsize{9}{9}\selectfont
    \setlength{\tabcolsep}{1mm}
    \centering
    \caption{Semantic Fusion Ablations}
    \vspace{-0.3cm}
    \begin{tabular}{cccccc}
        \toprule
        \multicolumn{6}{c}{YFCC26k GCD accuracy; higher is better} \\
        \midrule
        Fusion & Street & City  & Region & Country & Continent \\
        Blocks & 1 km   & 25 km & 200 km & 750 km  & 2500 km   \\
        \midrule
        3 & 17.8 & 31.5 & 45.1 & 64.3 & 79.3 \\
        2 & 14.6 & 31.6 & 45.4 & 62.5 & 77.7 \\
        1 & 13.9 & 30.7 & 44.0 & 61.2 & 76.8 \\
         None & 13.8 & 30.4 & 44.5 & 62.0 & 77.6 \\
        \midrule
        \multicolumn{6}{c}{GWS15k GCD accuracy; higher is better} \\
        \midrule
        Fusion & Street & City  & Region & Country & Continent \\
        Blocks & 1 km   & 25 km & 200 km & 750 km  & 2500 km   \\
        \midrule
        3 & 1.0 & 4.6 & 21.9 & 54.7 & 80.8 \\
        2 & 0.4 & 4.6 & 22.0 & 53.3 & 80.1 \\
        1 & 0.6 & 4.6 & 22.0 & 54.0 & 79.6 \\
         None & 0.6 & 4.6 & 23.0 & 54.6 & 81.2 \\
        \bottomrule
    \end{tabular}
    \label{tbl:ablation_sem}
\end{table}

Together, these ablations provide several insights into the hierarchical geographic representation. First, using geographic embeddings (i.e., a retrieval-based objective) consistently improves geo-localization performance over classification for both hierarchical and flat representations across the datasets in Tables~\ref{tbl:ablation_geo} and~\ref{tbl:ablation_geo2}. Second, GeoSURGE’s gains cannot be attributed solely to its hierarchy, as shown by the degraded performance in the second row of Table~\ref{tbl:ablation_geo}, where an architecture nearly identical to GeoSURGE retains the hierarchy but replaces retrieval with classification and therefore lacks geographic embeddings. Third, GeoSURGE's performance cannot be explained by geographic embeddings alone either, as demonstrated by the first row of Table~\ref{tbl:ablation_geo2}, which uses embeddings but collapses the hierarchy into a flat representation. In summary, both hierarchy and geographic embeddings are complementary and essential components of GeoSURGE’s core representation.

\label{sec:qualitative}

%% file: sec/5_conclusion.tex
\section{Conclusion}
\label{sec:conclusion}

In conclusion, GeoSURGE combines the strengths from classification and retrieval approaches, by learning a geographic representation that models the world as a hierarchy of geographic embeddings. Based on this geographic representation, GeoSURGE formulates the geo-localization problem as matching the visual representation of the query image with feature vectors from the geographic representation. In addition, GeoSURGE efficiently enriches the visual representation of the image by using latent cross-attention to fuse appearance features with semantic segmentation.

\begin{table}[t!] 
    \fontsize{9}{9}\selectfont
    \setlength{\tabcolsep}{1mm}
    \centering
    \caption{\underline{Hierarchical} Geographic Representation Ablations}
    \vspace{-0.3cm}
    \begin{tabular}{cccccc}
        \toprule
        \multicolumn{6}{c}{YFCC26k GCD accuracy; higher is better} \\
        \midrule
        Geographic & Street & City  & Region & Country & Continent \\
        Embeddings? & 1 km   & 25 km & 200 km & 750 km  & 2500 km   \\
        \midrule
        Yes & 17.8 & 31.5 & 45.1 & 64.3 & 79.3 \\
        \midrule
        No  & 15.1 & 31.5 & 45.3 & 63.0 & 78.2 \\
        \midrule
        \multicolumn{6}{c}{GWS15k GCD accuracy; higher is better} \\
        \midrule
        Geographic & Street & City  & Region & Country & Continent \\
        Embeddings? & 1 km   & 25 km & 200 km & 750 km  & 2500 km   \\
        \midrule
        Yes & 1.0 & 4.6 & 21.9 & 54.7 & 80.8 \\
        \midrule
        No  & 0.7 & 4.8 & 21.9 & 53.8 & 80.7 \\
        \bottomrule
    \end{tabular}
    \label{tbl:ablation_geo}
\end{table}

\begin{table}[t] 
    \fontsize{9}{9}\selectfont
    \setlength{\tabcolsep}{1mm}
    \centering
    \caption{\underline{Flat} Geographic Representation Ablations}
    \vspace{-0.3cm}
    \begin{tabular}{cccccc}
        \toprule
        \multicolumn{6}{c}{YFCC26k GCD accuracy; higher is better} \\
        \midrule
        Geographic & Street & City  & Region & Country & Continent \\
        Embeddings? & 1 km   & 25 km & 200 km & 750 km  & 2500 km   \\
        \midrule
        Yes & 14.2 & 30.2 & 43.5 & 61.2 & 76.9 \\
        \midrule
        No  & 14.0 & 30.1 & 43.2 & 61.0 & 77.4 \\
        \midrule
        \multicolumn{6}{c}{GWS15k GCD accuracy; higher is better} \\
        \midrule
        Geographic & Street & City  & Region & Country & Continent \\
        Embeddings? & 1 km   & 25 km & 200 km & 750 km  & 2500 km   \\
        \midrule
        Yes & 0.5 & 4.4 & 21.7 & 53.8 & 80.4 \\
        \midrule
        No  & 0.7 & 4.4 & 21.0 & 51.8 & 79.4 \\
        \bottomrule
    \end{tabular}
    \label{tbl:ablation_geo2}
\end{table}

GeoSURGE demonstrated new state-of-the-art results across 22 of 25 metrics on five benchmark datasets. The improvements over the most similar approaches, like GeoCLIP, underscore the importance of the geographic representation. Ablation studies showed that our geographic and visual representations were central to these performance improvements from GeoSURGE. While GeoSURGE’s strong performance suggests that explicit geographic modeling remains crucial for robust global geolocation, future advancements might be drawn by efficiently combining the benefits of LVLMs with structured geographic representations.

%% file: sec/X_suppl.tex
\clearpage
\setcounter{page}{1}
\maketitlesupplementary

\section{Ablation Studies for IM2GPS, IM2GPS3k, and YFCC4k}
\label{sec:supp_abaltions}

Below we provide results from hierarchy depth, semantic fusion, and geographic representation ablation studies for the remaining benchmark datasets which did not fit within the main body of the article. These include IM2GPS, IM2GPS3k, and YFCC4k.

\textbf{Hierarchy depth ablations} for IM2GPS, IM2GPS3k, and YFCC4k are provided in Table~\ref{tbl:supp_ablation_hier}. See Section~\ref{sec:eval_ablations} for details of the hierarchy depth ablations. We observed similar trends for the benchmarks presented here to the benchmarks presented in the main body of the article. Deeper hierarchies improve performance, stemming from smaller geocells with more precise locations as well as more diversity among geographical features at different scales.

\textbf{Semantic fusion ablations} for IM2GPS, IM2GPS3k, and YFCC4k are provided in Table~\ref{tbl:supp_ablation_sem}. See Section~\ref{sec:eval_ablations} for details of the semantic fusion ablations. We observed similar trends for the benchmarks presented here to the benchmarks presented in the main body of the article. Accuracy tends to increase with number of fusion blocks.

\textbf{Geographic representation ablations} for IM2GPS, IM2GPS3k, and YFCC4k are provided in Tables~\ref{tbl:supp_ablation_geo} and \ref{tbl:supp_ablation_geo2}. See Section~\ref{sec:eval_ablations} for details of the geographic representation ablations. We observed similar trends for the benchmarks presented here to the benchmarks presented in the main body of the article. In general, we observe that both hierarchy and geographic embeddings are complementary components of GeoSURGE’s core representation. Both are essential to achieving best overall performance.

\section{Frozen CLIP Ablations}
\label{sec:supp_frozenclip}

In GeoSURGE, all but the last CLIP Vision Transformer encoder blocks are kept frozen during training as described in Section~\ref{sec:approach_learning}. In this ablation study, we freeze all CLIP Vision Transformer encoder blocks to gauge the contribution of finetuning the last CLIP Vision Transformer encoder block for improved alignment between embeddings. All other variables are set to defaults and held constant across the GeoSURGE ablations. The results of the ablation are shown in Table \ref{tbl:frozen_clip}. While this ablation shows strong performance, fine-tuning the last CLIP layer to better merge with later layers gives further gains by improving the alignment between visual and geographic embeddings.

\begin{table}[t] 
    \fontsize{9}{9}\selectfont
    \setlength{\tabcolsep}{1mm}
    \centering
    \caption{Hierarchy Depth Ablations}
    \begin{tabular}{cccccc}
        \toprule
        \multicolumn{6}{c}{IM2GPS GCD accuracy; higher is better} \\
        \midrule
        Hierarchy & Street & City  & Region & Country & Continent \\
        Levels & 1 km   & 25 km & 200 km & 750 km  & 2500 km   \\
        \midrule
        7 & 27.0 & 54.4 & 70.0 & 84.4 & 93.2 \\
        5 & 19.4 & 48.9 & 67.9 & 82.3 & 92.4 \\
        3 & 18.1 & 48.5 & 66.7 & 82.7 & 92.4 \\
        1 & 18.1 & 45.6 & 67.1 & 84.0 & 92.8 \\
        \midrule
        \multicolumn{6}{c}{IM2GPS3k GCD accuracy; higher is better} \\
        \midrule
        Hierarchy & Street & City  & Region & Country & Continent \\
        Levels & 1 km   & 25 km & 200 km & 750 km  & 2500 km   \\
        \midrule
        7 & 17.2 & 42.5 & 58.1 & 74.6 & 87.6 \\
        5 & 13.5 & 40.2 & 56.7 & 74.1 & 86.9 \\
        3 & 12.4 & 39.0 & 56.5 & 73.6 & 86.5 \\
        1 & 11.0 & 37.8 & 56.5 & 73.2 & 86.0 \\
        \midrule
        \multicolumn{6}{c}{YFCC4k GCD accuracy; higher is better} \\
        \midrule
        Hierarchy & Street & City  & Region & Country & Continent \\
        Levels & 1 km   & 25 km & 200 km & 750 km  & 2500 km   \\
        \midrule
        7 & 19.9 & 33.6 & 48.7 & 67.4 & 82.0 \\
        5 & 12.2 & 31.8 & 48.2 & 66.3 & 80.9 \\
        3 & 10.7 & 31.0 & 46.9 & 65.4 & 81.1 \\
        1 & 8.4 & 29.5 & 47.1 & 65.3 & 80.9 \\
        \bottomrule
    \end{tabular}
    \label{tbl:supp_ablation_hier}
\end{table}

\begin{table}[t] 
    \fontsize{9}{9}\selectfont
    \setlength{\tabcolsep}{1mm}
    \centering
    \caption{Semantic Fusion Ablations}
    \begin{tabular}{cccccc}
        \toprule
        \multicolumn{6}{c}{IM2GPS GCD accuracy; higher is better} \\
        \midrule
        Fusion & Street & City  & Region & Country & Continent \\
        Blocks & 1 km   & 25 km & 200 km & 750 km  & 2500 km   \\
        \midrule
        3 & 27.0 & 54.4 & 70.0 & 84.4 & 93.2 \\
        2 & 22.4 & 49.8 & 68.8 & 83.1 & 93.2 \\
        1 & 21.5 & 50.2 & 67.1 & 83.5 & 91.1 \\
        None & 19.8 & 49.8 & 68.3 & 83.5 & 93.2 \\
        \midrule
        \multicolumn{6}{c}{IM2GPS3k GCD accuracy; higher is better} \\
        \midrule
        Fusion & Street & City  & Region & Country & Continent \\
        Blocks & 1 km   & 25 km & 200 km & 750 km  & 2500 km   \\
        \midrule
        3 & 17.2 & 42.5 & 58.1 & 74.6 & 87.6 \\
        2 & 16.1 & 42.0 & 58.2 & 75.0 & 86.8 \\
        1 & 16.8 & 41.9 & 57.2 & 73.6 & 86.1 \\
        None & 15.3 & 41.5 & 57.0 & 74.0 & 86.3 \\
        \midrule
        \multicolumn{6}{c}{YFCC4k GCD accuracy; higher is better} \\
        \midrule
        Fusion & Street & City  & Region & Country & Continent \\
        Blocks & 1 km   & 25 km & 200 km & 750 km  & 2500 km   \\
        \midrule
        3 & 19.9 & 33.6 & 48.7 & 67.4 & 82.0 \\
        2 & 15.1 & 33.6 & 48.9 & 66.8 & 81.8 \\
        1 & 15.8 & 33.6 & 48.7 & 66.0 & 81.0 \\
        None & 14.4 & 31.7 & 47.8 & 65.5 & 80.8 \\
        \bottomrule
    \end{tabular}
    \label{tbl:supp_ablation_sem}
\end{table}

\begin{table}[t!] 
    \fontsize{9}{9}\selectfont
    \setlength{\tabcolsep}{1mm}
    \centering
    \caption{\underline{Hierarchical} Geographic Representation Ablations}
    \begin{tabular}{cccccc}
        \toprule
        \multicolumn{6}{c}{IM2GPS GCD accuracy; higher is better} \\
        \midrule
        Geographic & Street & City  & Region & Country & Continent \\
        Embeddings? & 1 km   & 25 km & 200 km & 750 km  & 2500 km   \\
        \midrule
        Yes & 27.0 & 54.4 & 70.0 & 84.4 & 93.2 \\
        \midrule
        No  & 22.8 & 54.0 & 71.3 & 83.5 & 92.8 \\
        \midrule
        \multicolumn{6}{c}{IM2GPS3k GCD accuracy; higher is better} \\
        \midrule
        Geographic & Street & City  & Region & Country & Continent \\
        Embeddings? & 1 km   & 25 km & 200 km & 750 km  & 2500 km   \\
        \midrule
        Yes & 17.2 & 42.5 & 58.1 & 74.6 & 87.6 \\
        \midrule
        No  & 17.4 & 43.5 & 59.1 & 75.6 & 87.6 \\
        \midrule
        \multicolumn{6}{c}{YFCC4k GCD accuracy; higher is better} \\
        \midrule
        Geographic & Street & City  & Region & Country & Continent \\
        Embeddings? & 1 km   & 25 km & 200 km & 750 km  & 2500 km   \\
        \midrule
        Yes & 19.9 & 33.7 & 49.4 & 67.4 & 82.0 \\
        \midrule
        No  & 15.7 & 33.6 & 48.7 & 67.4 & 81.5 \\
        \bottomrule
    \end{tabular}
    \label{tbl:supp_ablation_geo}
\end{table}

\begin{table}[t] 
    \fontsize{9}{9}\selectfont
    \setlength{\tabcolsep}{1mm}
    \centering
    \caption{\underline{Flat} Geographic Representation Ablations}
    \begin{tabular}{cccccc}
        \toprule
        \multicolumn{6}{c}{IM2GPS GCD accuracy; higher is better} \\
        \midrule
        Geographic & Street & City  & Region & Country & Continent \\
        Embeddings? & 1 km   & 25 km & 200 km & 750 km  & 2500 km   \\
        \midrule
        Yes & 22.4 & 51.5 & 67.5 & 83.4 & 91.1 \\
        \midrule
        No  & 22.8 & 53.2 & 67.1 & 84.0 & 93.2 \\
        \midrule
        \multicolumn{6}{c}{IM2GPS3k GCD accuracy; higher is better} \\
        \midrule
        Geographic & Street & City  & Region & Country & Continent \\
        Embeddings? & 1 km   & 25 km & 200 km & 750 km  & 2500 km   \\
        \midrule
        Yes & 15.9 & 41.8 & 56.8 & 73.2 & 86.5 \\
        \midrule
        No  & 15.9 & 41.4 & 56.9 & 73.4 & 85.5 \\
        \midrule
        \multicolumn{6}{c}{YFCC4k GCD accuracy; higher is better} \\
        \midrule
        Geographic & Street & City  & Region & Country & Continent \\
        Embeddings? & 1 km   & 25 km & 200 km & 750 km  & 2500 km   \\
        \midrule
        Yes & 15.2 & 32.9 & 47.4 & 65.6 & 80.5 \\
        \midrule
        No  & 14.9 & 32.6 & 47.5 & 64.9 & 80.5 \\
        \bottomrule
    \end{tabular}
    \label{tbl:supp_ablation_geo2}
\end{table}

\begin{table}[t!] 
    \fontsize{9}{9}\selectfont
    \setlength{\tabcolsep}{1mm}
    \centering
    \caption{Frozen CLIP Ablations}
    \begin{tabular}{cccccc}
        \toprule
        \multicolumn{6}{c}{IM2GPS GCD accuracy; higher is better} \\
        \midrule
        Finetune & Street & City  & Region & Country & Continent \\
        CLIP? & 1 km   & 25 km & 200 km & 750 km  & 2500 km   \\
        \midrule
        Yes & 27.0 & 54.4 & 70.0 & 84.4 & 93.2 \\
        \midrule
        No  & 24.1      & 52.3     & 70.5      & 84.0       & 93.2 \\
        \midrule
        \multicolumn{6}{c}{IM2GPS3k GCD accuracy; higher is better} \\
        \midrule
        Finetune & Street & City  & Region & Country & Continent \\
        CLIP? & 1 km   & 25 km & 200 km & 750 km  & 2500 km   \\
        \midrule
        Yes & 17.2 & 42.5 & 58.1 & 74.6 & 87.6 \\
        \midrule
        No  & 16.7      & 41.6     & 57.3      & 74.4       & 86.4 \\
        \midrule
        \multicolumn{6}{c}{YFCC4k GCD accuracy; higher is better} \\
        \midrule
        Finetune & Street & City  & Region & Country & Continent \\
        CLIP? & 1 km   & 25 km & 200 km & 750 km  & 2500 km   \\
        \midrule
        Yes & 19.9 & 33.6 & 48.7 & 67.4 & 82.0 \\
        \midrule
        No  & 19.3      & 33.5     & 47.3      & 65.5       & 80.9 \\
        \midrule
        \multicolumn{6}{c}{YFCC26k GCD accuracy; higher is better} \\
        \midrule
        Finetune & Street & City  & Region & Country & Continent \\
        CLIP? & 1 km   & 25 km & 200 km & 750 km  & 2500 km   \\
        \midrule
        Yes & 17.8 & 31.5 & 45.1 & 64.3 & 79.3 \\
        \midrule
        No  & 15.7 & 29.5 & 43.4 & 61.3 & 77.2 \\
        \midrule
        \multicolumn{6}{c}{GWS15k GCD accuracy; higher is better} \\
        \midrule
        Finetune & Street & City  & Region & Country & Continent \\
        CLIP? & 1 km   & 25 km & 200 km & 750 km  & 2500 km   \\
        \midrule
        Yes & 1.0 & 4.6 & 21.9 & 54.7 & 80.8 \\
        \midrule
        No  & 0.9 & 4.8 & 22.9 & 54.0 & 81.1 \\
        \bottomrule
    \end{tabular}
    \label{tbl:frozen_clip}
\end{table}

%% file: main.bib
@String(ECCV= {Eur. Conf. Comput. Vis.})

@String(ECCV  = {ECCV})

@article{jia2024g3,
  title={G3: an effective and adaptive framework for worldwide geolocalization using large multi-modality models},
  author={Jia, Pengyue and Liu, Yiding and Li, Xiaopeng and Zhao, Xiangyu and Wang, Yuhao and Du, Yantong and Han, Xiao and Wei, Xuetao and Wang, Shuaiqiang and Yin, Dawei},
  journal={Advances in Neural Information Processing Systems},
  volume={37},
  pages={53198--53221},
  year={2024}
}

@inproceedings{haas2024pigeon,
  title={Pigeon: Predicting image geolocations},
  author={Haas, Lukas and Skreta, Michal and Alberti, Silas and Finn, Chelsea},
  booktitle={Proceedings of the IEEE/CVF Conference on Computer Vision and Pattern Recognition},
  pages={12893--12902},
  year={2024}
}

@article{vivanco2024geoclip,
  title={Geoclip: Clip-inspired alignment between locations and images for effective worldwide geo-localization},
  author={Vivanco Cepeda, Vicente and Nayak, Gaurav Kumar and Shah, Mubarak},
  journal={Advances in Neural Information Processing Systems},
  volume={36},
  year={2024}
}

@inproceedings{zhou2024img2loc,
  title={Img2Loc: Revisiting Image Geolocalization using Multi-modality Foundation Models and Image-based Retrieval-Augmented Generation},
  author={Zhou, Zhongliang and Zhang, Jielu and Guan, Zihan and Hu, Mengxuan and Lao, Ni and Mu, Lan and Li, Sheng and Mai, Gengchen},
  booktitle={Proceedings of the 47th International ACM SIGIR Conference on Research and Development in Information Retrieval},
  pages={2749--2754},
  year={2024}
}

@inproceedings{clark2023we,
  title={Where we are and what we're looking at: Query based worldwide image geo-localization using hierarchies and scenes},
  author={Clark, Brandon and Kerrigan, Alec and Kulkarni, Parth Parag and Cepeda, Vicente Vivanco and Shah, Mubarak},
  booktitle={Proceedings of the IEEE/CVF Conference on Computer Vision and Pattern Recognition},
  pages={23182--23190},
  year={2023}
}

@inproceedings{theiner2022interpretable,
  title={Interpretable semantic photo geolocation},
  author={Theiner, Jonas and M{\"u}ller-Budack, Eric and Ewerth, Ralph},
  booktitle={Proceedings of the IEEE/CVF Winter Conference on Applications of Computer Vision},
  pages={750--760},
  year={2022}
}

@inproceedings{muller2018geolocation,
  title={Geolocation estimation of photos using a hierarchical model and scene classification},
  author={Muller-Budack, Eric and Pustu-Iren, Kader and Ewerth, Ralph},
  booktitle={Proceedings of the European conference on computer vision (ECCV)},
  pages={563--579},
  year={2018}
}

@inproceedings{weyand2016planet,
  title={Planet-photo geolocation with convolutional neural networks},
  author={Weyand, Tobias and Kostrikov, Ilya and Philbin, James},
  booktitle={Computer Vision--ECCV 2016: 14th European Conference, Amsterdam, The Netherlands, October 11-14, 2016, Proceedings, Part VIII 14},
  pages={37--55},
  year={2016},
  organization={Springer}
}

@inproceedings{seo2018cplanet,
  title={Cplanet: Enhancing image geolocalization by combinatorial partitioning of maps},
  author={Seo, Paul Hongsuck and Weyand, Tobias and Sim, Jack and Han, Bohyung},
  booktitle={Proceedings of the European Conference on Computer Vision (ECCV)},
  pages={536--551},
  year={2018}
}

@inproceedings{izbicki2020exploiting,
  title={Exploiting the earth’s spherical geometry to geolocate images},
  author={Izbicki, Mike and Papalexakis, Evangelos E and Tsotras, Vassilis J},
  booktitle={Machine Learning and Knowledge Discovery in Databases: European Conference, ECML PKDD 2019, W{\"u}rzburg, Germany, September 16--20, 2019, Proceedings, Part II},
  pages={3--19},
  year={2020},
  organization={Springer}
}

@inproceedings{kordopatis2021leveraging,
  title={Leveraging efficientnet and contrastive learning for accurate global-scale location estimation},
  author={Kordopatis-Zilos, Giorgos and Galopoulos, Panagiotis and Papadopoulos, Symeon and Kompatsiaris, Ioannis},
  booktitle={Proceedings of the 2021 International Conference on Multimedia Retrieval},
  pages={155--163},
  year={2021}
}

@inproceedings{pramanick2022world,
  title={Where in the world is this image? transformer-based geo-localization in the wild},
  author={Pramanick, Shraman and Nowara, Ewa M and Gleason, Joshua and Castillo, Carlos D and Chellappa, Rama},
  booktitle={European Conference on Computer Vision},
  pages={196--215},
  year={2022},
  organization={Springer}
}

@inproceedings{astruc2024openstreetview,
  title={OpenStreetView-5M: The Many Roads to Global Visual Geolocation},
  author={Astruc, Guillaume and Dufour, Nicolas and Siglidis, Ioannis and Aronssohn, Constantin and Bouia, Nacim and Fu, Stephanie and Loiseau, Romain and Nguyen, Van Nguyen and Raude, Charles and Vincent, Elliot and others},
  booktitle={Proceedings of the IEEE/CVF Conference on Computer Vision and Pattern Recognition},
  pages={21967--21977},
  year={2024}
}

@inproceedings{luo2022g3,
  title={G3: Geolocation via Guidebook Grounding},
  author={Luo, Grace and Biamby, Giscard and Darrell, Trevor and Fried, Daniel and Rohrbach, Anna},
  booktitle={Findings of the Association for Computational Linguistics: EMNLP 2022},
  pages={5841--5853},
  year={2022}
}

@inproceedings{wu2022im2city,
  title={IM2City: image geo-localization via multi-modal learning},
  author={Wu, Meiliu and Huang, Qunying},
  booktitle={Proceedings of the 5th ACM SIGSPATIAL International Workshop on AI for Geographic Knowledge Discovery},
  pages={50--61},
  year={2022}
}

@article{hays2015large,
  title={Large-scale image geolocalization},
  author={Hays, James and Efros, Alexei A},
  journal={Multimodal location estimation of videos and images},
  pages={41--62},
  year={2015},
  publisher={Springer}
}

@inproceedings{radford2021learning,
  title={Learning transferable visual models from natural language supervision},
  author={Radford, Alec and Kim, Jong Wook and Hallacy, Chris and Ramesh, Aditya and Goh, Gabriel and Agarwal, Sandhini and Sastry, Girish and Askell, Amanda and Mishkin, Pamela and Clark, Jack and others},
  booktitle={International conference on machine learning},
  pages={8748--8763},
  year={2021},
  organization={PMLR}
}

@inproceedings{jain2023oneformer,
  title={Oneformer: One transformer to rule universal image segmentation},
  author={Jain, Jitesh and Li, Jiachen and Chiu, Mang Tik and Hassani, Ali and Orlov, Nikita and Shi, Humphrey},
  booktitle={Proceedings of the IEEE/CVF Conference on Computer Vision and Pattern Recognition},
  pages={2989--2998},
  year={2023}
}

@article{zhou2019semantic,
  title={Semantic understanding of scenes through the ade20k dataset},
  author={Zhou, Bolei and Zhao, Hang and Puig, Xavier and Xiao, Tete and Fidler, Sanja and Barriuso, Adela and Torralba, Antonio},
  journal={International Journal of Computer Vision},
  volume={127},
  number={3},
  pages={302--321},
  year={2019},
  publisher={Springer}
}

@inproceedings{he2016deep,
  title={Deep residual learning for image recognition},
  author={He, Kaiming and Zhang, Xiangyu and Ren, Shaoqing and Sun, Jian},
  booktitle={Proceedings of the IEEE conference on computer vision and pattern recognition},
  pages={770--778},
  year={2016}
}

@article{oord2018representation,
  title={Representation learning with contrastive predictive coding},
  author={Oord, Aaron van den and Li, Yazhe and Vinyals, Oriol},
  journal={arXiv preprint arXiv:1807.03748},
  year={2018}
}

@inproceedings{hays2008im2gps,
  title={Im2gps: estimating geographic information from a single image},
  author={Hays, James and Efros, Alexei A},
  booktitle={2008 ieee conference on computer vision and pattern recognition},
  pages={1--8},
  year={2008},
  organization={IEEE}
}

@inproceedings{vo2017revisiting,
  title={Revisiting im2gps in the deep learning era},
  author={Vo, Nam and Jacobs, Nathan and Hays, James},
  booktitle={Proceedings of the IEEE international conference on computer vision},
  pages={2621--2630},
  year={2017}
}

@article{masone2021survey,
  title={A survey on deep visual place recognition},
  author={Masone, Carlo and Caputo, Barbara},
  journal={IEEE Access},
  volume={9},
  pages={19516--19547},
  year={2021},
  publisher={IEEE}
}

@inproceedings{lin2013cross,
  title={Cross-view image geolocalization},
  author={Lin, Tsung-Yi and Belongie, Serge and Hays, James},
  booktitle={Proceedings of the IEEE Conference on Computer Vision and Pattern Recognition},
  pages={891--898},
  year={2013}
}

@article{guo2025deepseek,
  title={Deepseek-r1: Incentivizing reasoning capability in llms via reinforcement learning},
  author={Guo, Daya and Yang, Dejian and Zhang, Haowei and Song, Junxiao and Zhang, Ruoyu and Xu, Runxin and Zhu, Qihao and Ma, Shirong and Wang, Peiyi and Bi, Xiao and others},
  journal={arXiv preprint arXiv:2501.12948},
  year={2025}
}

@inproceedings{dufour2025around,
  title={Around the world in 80 timesteps: A generative approach to global visual geolocation},
  author={Dufour, Nicolas and Kalogeiton, Vicky and Picard, David and Landrieu, Loic},
  booktitle={Proceedings of the Computer Vision and Pattern Recognition Conference},
  pages={23016--23026},
  year={2025}
}

@inproceedings{li2024georeasoner,
  title={GeoReasoner: Geo-localization with reasoning in street views using a large vision-language model},
  author={Li, Ling and Ye, Yu and Jiang, Bingchuan and Zeng, Wei},
  booktitle={Proceedings of the 41st International Conference on Machine Learning},
  pages={29222--29233},
  year={2024}
}

@inproceedings{campos2026gaea,
  title={Gaea: A geolocation aware conversational assistant},
  author={Campos, Ron and Vayani, Ashmal and Kulkarni, Parth Parag and Gupta, Rohit and Zafar, Aizan and Dutta, Aritra and Shah, Mubarak},
  booktitle={Proceedings of the IEEE/CVF Winter Conference on Applications of Computer Vision},
  pages={5236--5246},
  year={2026}
}
